\documentclass[12pt,a4paper,twoside, twocolumn]{article}
\usepackage{dhasa}
\usepackage{hyperref}
\begin{document}
\title*{Training Cross-Lingual embeddings for Setswana and Sepedi}
\textit{Makgatho, Mack}\\
\textit{Dept. of Computer Science, University of Pretoria} \\
\textit{mack.letladi1@gmail.com}\\
\\
\textit{Marivate, Vukosi}\\
\textit{Dept. of Computer Science, University of Pretoria} \\
\textit{vukosi.marivate@cs.up.ac.za}\\
\\
\textit{Sefara, Tshephisho}\\
\textit{Council for Scientific and Industrial Research} \\
\textit{tsefara@csir.co.za}\\
\\
\textit{Wagner, Valencia}\\
\textit{Sol Plaatje University} \\
\textit{valencia.wagner@spu.ac.za} 

\section*{Abstract}
African languages still lag in the advances of Natural Language Processing techniques, one reason being the lack of representative data, having a technique that can transfer information between languages can help mitigate against the lack of data problem. This paper trains Setswana and Sepedi monolingual word vectors and uses VecMap to create cross-lingual embeddings for Setswana-Sepedi in order to do a cross-lingual transfer.

Word embeddings are word vectors that represent words as continuous floating numbers where semantically similar words are mapped to nearby points in n-dimensional space. The idea of word embeddings is based on the distribution hypothesis that states, semantically similar words are distributed in similar contexts (Harris, 1954). 

Cross-lingual embeddings leverages monolingual embeddings by learning a shared vector space for two separately trained monolingual vectors such that words with similar meaning are represented by similar vectors. In this paper, we investigate cross-lingual embeddings for Setswana-Sepedi monolingual word vector. We use the unsupervised cross lingual embeddings in VecMap to train the Setswana-Sepedi cross-language word embeddings. We evaluate the quality of the Setswana-Sepedi cross-lingual word representation using a semantic evaluation task. For the semantic similarity task, we translated the WordSim and SimLex tasks into Setswana and Sepedi. We release this dataset as part of this work for other researchers. We evaluate the intrinsic quality of the embeddings to determine if there is improvement in the semantic representation of the word embeddings.


 



Keywords: cross-lingual embeddings, word embeddings, intrinsic evaluation

\section{Introduction}

Many African languages have insufficient language resources (data, tools, people) \citep{abbott2019benchmarking,martinus2019focus,nekoto2020participatory,sefara2021transformer} and fall into the classification of low resource languages \citep{ranathunga2021neural} in the Natural Language Processing (NLP) field. This lack of resources makes it harder to capitalise on the recent advances in many NLP downstream tasks such as Neural Machine Translation \citep{cho2014learning}, Large Language Models \citep{devlin2018bert,howard2018universal}, Q\&A systems \citep{kwiatkowski2019natural}, etc. There may be more downstream approaches to deal with some of these challenges such as Transfer Learning \citep{ruder2019transfer}, Data Augmentation \citep{marivate2020improving}, Multilingual Models \citep{hedderich2020transfer}, etc. Additionally, the lack of research attention to existing NLP techniques results in difficulties finding a benchmark \citep{abbott2019benchmarking}. In this work, we focus on word representations through word embeddings and how we can leverage one language to assist in the representation of another related language. These embeddings can then be used to develop tools for other downstream tasks.

Word Embeddings are a mathematical technique to learn general language vector representations from a large amount of unlabelled text using co-occurring statistics. In recent years, monolingual word embeddings techniques are increasingly becoming an important resource in NLP. Word embeddings are widely used in NLP problems such as sentiment analysis \citep{socher2013recursive}, named-entity-recognition \citep{guo2014revisiting}, parts-of-speech tagging, and document retrieval. Word2Vec is a vector training model proposed by \cite{mikolov2013distributed}. Word2Vec produces a low-dimensional real-value vector representing the meaning of a word. The word vector represents grammatical and semantic properties, which results in words with similar semantic relations being close to each other. The word vector representation method incorporates the semantic relationship between words which is not possible through representations such as Bag-Of-Words of TFIDF. Word embeddings are better than both methods because they map all the words in a language into a vector space of a given dimension, the words are converted into vectors and allow multiple linear operations and have the property of preserving analogies \citep{mikolov2013distributed,inproceedings}.

Cross-lingual word embeddings have been receiving more and more attention from the NLP community, mainly because it has provided a path to effectively align two disjoint monolingual embeddings with no bilingual dictionary for unsupervised techniques or no more than a small bilingual dictionary for supervised techniques \citep{lample2018word, artetxe2020cross}. Cross-lingual techniques also enable knowledge transfer between languages with rich resources and low resources. For languages lacking bilingual parallel corpus with other languages, cross-lingual embeddings can be utilised to train high-quality cross-lingual embeddings \citep{lample2018word}. This can aid in accelerating the progress of applying NLP to low-resourced languages. \cite{artetxe2018robust} created the cross-lingual unsupervised or supervised word embedding (VecMap library) approach for training cross-lingual word embedding models. The approaches can be used to construct cross-language word vectors with or without a bilingual dictionary.

The majority of South African languages lag bilingual parallel corpus with other languages. In this work, we aim to investigate how cross-lingual embeddings could be used to improve the state of one or both languages. We used data (corpus) from different domains to train Word2Vec and fastText \citep{bojanowski2016enriching} monolingual embeddings. When using VecMap, the two embeddings are aligned. VecMap requires two monolingual word vectors from source and target \citep{artetxe2018robust}. To evaluate the effectiveness of the cross-lingual embedding for Setswana and Sepedi, we use intrinsic evaluation \citep{bakarov2018survey} through Setswana and Sepedi versions of WordSim \citep{finkelstein2001placing} and Simlex \citep{hill2015Simlex}. This is following on an approach that has been used for Yoruba and Twi \citep{alabi2019massive}. We also release the dataset for this benchmark of human semantic similarity task.

This paper is structured as follows; the next section is a review of related work that is done on cross-lingual word vectors. Followed by data collection in Section~\ref{sec:data}. Section~\ref{sce:train} discusses methodology followed to train cross-lingual word vectors using VecMap. The evaluation of the word vectors is discussed in Section~\ref{sec:eva}. Section~\ref{sec:res} explains the results while Section~\ref{sec:disc} discusses the findings and finally, conclusions and future work can be found in Section~\ref{sec:con}. 

\section{Background and Related Work}

Cross-lingual word embeddings (CLWEs) are becoming popular in NLP for two reasons: Cross-lingual word embeddings can transfer knowledge from rich-resourced languages to low-resourced; The technique can also infer the semantics of words in a multiple language environment. \cite{conneau2018word} show that word embeddings spaces can be aligned without any cross-lingual supervision. The alignment is based on solely unaligned datasets of each language. Using adversarial training, they were able to initialise a linear mapping between a source and a target space, which they use to create a synthetic parallel dictionary. First, they propose a simple criterion that is used as an unsupervised validation matric. Second, they propose the similarity measure cross-domain similarity local scaling (CSLS), which mitigates the hubness problem and increases the word translation accuracy. The hubness problem is defined by \cite{dinu2015improving} as:\begin{quote}
    "neighbourhoods of the mapped elements are strongly polluted by hubs, vectors that tend to be near a high proportion of items, pushing their correct labels down the neighbour list."
\end{quote} In the work done by \cite{adams2017cross}, the research looked at applying CLWEs to Yongning Na, a Sino-Tibetan language. The research focused on determining if the quality of CLWEs depends on having large amounts of data in multiple languages and if initialising the parameters of neural network language models (NMLM) can improve language modelling in a low-resourced context. The research scaled down the available monolingual data of the target language to about 1000 sentences. The quality of intrinsic embedding was assessed by taking into consideration correlation with human judgement on the WordSim353 \citep{finkelstein2001placing} test set. They went further to perform language modelling experiments by initialising the parameters for long short-term memory (LSTM) \citep{Hochreiter1997LongSM} by training across different language pairs. The research showed that CLWEs are resilient even when target language training data is scaled-down and that initialisation of NMLM parameters leads to good performance.

\cite{artetxe2019massively} introduced an architecture that can be used to learn multilingual sentence representations for more than 90 languages. The languages belonged to 30 different families. The research used a single BiLSTM encoder with a shared Byte Pair Encoding (BPE) vocabulary coupled with an auxiliary decoder and trained on parallel corpora. They learn a classifier using English annotated data only and transfer it to any language without modification. The research mainly focused on vector representations of sentences that are general for the input language and the NLP task.

\cite{alabi2019massive} worked on massive vs. curated embeddings for low-resourced languages: the case of Yorùbá and Twi. Authors compare two types of word embeddings obtained from curated corpora and a language-dependent processing. They move further to collect high quality and noisy data for the two languages. They quantify that improvements that is based on the quality of data and not only on the amount of data. In their experiments, they use different architectures to learn word representations both from characters and surface forms. They evaluate multilingual BERT on a down stream task, specifically named entity recognition and WordSim-353 word pairs dataset.

\cite{ijcai2018-566} investigates a cross-lingual knowledge transfer technique to improve the semantic representation of low-resourced languages and improving low resource named-entity recognition. In their research, neural networks are used to do knowledge transfer from high resource language using bilingual lexicons to improve low resource word representation. They automatically learn semantic projections using a lexicon extension strategy that is designed to address out-of lexicon problem. Finally, they regard word-level entity type distribution features as an external language independent knowledge and incorporate them into their neural architecture. The experiment is done on two low resource languages (Dutch and Spanish) to demonstrate the effectiveness of these additional semantic representations.

\cite{banerjee2021crosslingual} show that initialising the embedding layer of Unsupervised Neural Machine Translation (UNMT) models with cross-lingual embeddings shows significant improvements in BLEU score. Authors show that freezing the embedding layer weights lead to better gains compared to updating the embedding layer weights during training. They experimented using Denoising Autoencoder (DAE) and Masked Sequence to Sequence (MASS) for three different unrelated language pairs (for English-Hindi, English-Bengali, and English-Gujarati). The analysis shows the importance of using cross-lingual embedding as compared to other techniques.


The literature shows that there is a substantial amount of work done on cross-lingual transfer and empirical proof that the method improves the performance of models. The literature does not relay solely on intrinsic evaluation but the solutions are applied to some downstream tasks. In the next section, we detail the data used for conducting experiments. 

\section{Data collection}\label{sec:data}

Training data is very important for implementing powerful and accurate models, and clean training data can make a difference between a good and great model. The data needs to be very imperative because the quality of the alignment depends on the quality of the monolingual embeddings, i.e. data used to create the initial monolingual embeddings before mapping.

We use data collected from different domains for training word vectors:
\begin{itemize}
    \item \textbf{JW300 bible} \citep{agic-vulic-2019-jw300}: A biblical-domain data set containing parallel corpus for low-resourced languages. 
    \item \textbf{Wikipedia}
    \item \textbf{National Centre for Human Language Technology (NCHLT) text corpus} \citep{eiselen2014developing}: The dataset contains clean textual data in Sepedi and Setswana. The data set was constructed by harvesting existing data such as online publications, online news, web crawling and crowd-sourcing.
    \item \textbf{SABC News Data in Setswana and Sepedi} \citep{marivate2020investigating, marivate_vukosi_2020_3668495}: The data set contains news titles collected from online social media.
\end{itemize}
National Centre for Human Language Technology (NCHLT) data is used for training monolingual word vectors. For preprocessing, we changed all words to lowercase, removing brackets, digits, punctuations, and white spaces.


 \begin{table}[t]
 \caption{Corpus size for the Setswana and Sepedi Datasets}\label{tab3}
 \begin{tabular}{llll} 
 & Sepedi & Setswana  \\ 
 Number of tokens: &2133972 &3000682 \\
 Unique words: &93461 &107606 \\

 \end{tabular}
 \end{table}

In this section we dealt with how we collected the data used to training our monolingual embeddings for both languages and what approach we took to pre-process the data before training the models. In the next section we discuss the approach taken to train the monolingual embeddings and how VecMap was used to training the cross-lingual embeddings.

\section{Training monolingual and cross-lingual embeddings (VecMap)}\label{sce:train}

In this section, we present the methods (frameworks) used to train monolingual and cross-lingual embeddings. We describe the parameters used to train word2Vec and fastText embeddings. We also look into VecMap, the framework that we used to align monolingual embeddings.

CLWEs have proved to perform very well for low-resourced languages. The main idea is to do a cross-lingual transfer from the source language to the target, such that we have a single representation for a pair of languages where semantically similar words are closer to one another. In order to use VecMap two monolingual embeddings are required, we train fastText and word2Vec vectors. We use the following parameters for fastText and word2Vec in Table~\ref{tab2}. The definition of the parameters are as follows: skipGram - training method, dim - size of word vectors, minCount - minimal number of word occurrences, ws - size of the context window, and epoch - number of epochs or iterations.

 \begin{table}[t]
 \caption{Parameters for FastText and Word2Vec}\label{tab2}
 \begin{tabular}{ll} 
  Parameter & Value \\
  skip-gram	& true \\
  dim	& 300 \\
  minCount & 1 \\
  ws & 4 \\
  epoch & 100 \\
 \end{tabular}
 \end{table}
 
\subsection{Word2Vec}

The word2Vec \citep{mikolov2013distributed} algorithm is a two-layer neural network that vectorises words to processes text. The algorithm takes as input a text corpus and returns feature vectors that represent words in that corpus as a set of vectors. Word2Vec trains words against neighbouring words based on a window size context. It trains the words using two methods: skip-gram or continuous bag of words (CBOW), skip-gram uses a word to predict a target context and CBOW uses context to predict a target word. The experiment uses skip-gram to train monolingual embeddings. We use word vectors that were trained using Word2Vec. These correspond to monolingual embeddings of dimension 300 trained on Sepedi and Setswana corpora. 

\subsection{FastText}

FastText \citep{bojanowski2016enriching} is a supervised prediction-based technique based on the word2Vec family of algorithms \citep{mikolov2013distributed}. It predicts tags through context and represents each word as an \textit{n}-gram of characters, instead of learning vectors for words directly. The fastText model has three layers: input layer, hidden layer, and output layer. Input is a number of words and their \textit{n}-gram features, these features are used to represent a single document. The hidden layer is the superimposed average of multiple feature vectors. The hidden layer solves the maximum likelihood function, then constructs a Huffman tree according to the weights and model parameters of each category, and uses the Huffman tree as the output.

\subsection{VecMap}
VecMap \citep{artetxe2020cross} is an open-source framework to learn CLWEs written in Python. There are two techniques to do cross-lingual embeddings with VecMap, supervised (recommended if you have a large training dictionary) and unsupervised (recommended if you have no seed dictionary and do not want to rely on identical words). In this work, we align word embedding using VecMap\footnote{\url{https://github.com/artetxem/VecMap}}. The approach is fully unsupervised. The steps we followed to build our cross-lingual word embeddings model are motivated by the authors of VecMap \cite{artetxe2020cross}. The assumption is that we have a monolingual corpus for source and target languages. The word representations is learned independently for each language (monolingual embeddings for each language), and then mapped to a common vector space.

In this section, we presented word2Vec, fastText and VecMap. We also described the parameters used to train word2Vec and fastText embeddings. In the next section, we present experimental results and perform some analyses.

\section{Evaluation}\label{sec:eva}


We evaluate the quality of Setswana and Sepedi word vector representations on two different benchmarks Simlex and WordSim. The datasets (Simlex and WordSim) contain pairs of Setswana and Sepedi words that have been assigned similarity ratings by humans. They give a similarity score between a pair of words corresponding to their relatedness. Cosine similarity is used to collect a score from the model in order to check how close the score is to the human score, we use Spearman to measure correlation. Spearman index measure the dependence of two variables, the correlation of two statistical variables is evaluated using monotonic equation. We manually translate the WordSim and Simlex word pairs dataset from English into Setswana and Sepedi. We are releasing a dataset of Setswana and Sepedi translated WordSim and Simlex as part of this project at \url{https://github.com/dsfsi/embedding-eval-data} 
and archived on Zenodo at \url{https://zenodo.org/record/5673974}.

\subsection{Results}\label{sec:res}


This section presents the results of the experiments conducted to show the efficiency of the proposed technique with a couple of experiments. We first present the monolingual evaluation task for word2Vec and fastText and then present the cross-lingual evaluation task for Setswana and Sepedi. The evaluations of cross-lingual evaluation task is based on two embedding methods fastText and word2Vec.

In Table~\ref{tabfast} and Table~\ref{tabword}, we show the Spearman’s correlation for word vectors trained on fastText and word2vec. The correlation scores calculate the similarity between word vectors. Table~\ref{tabcrossw} and Table~\ref{tabcrossf} scores are obtained from using Setswana and Sepedi monolingual vectors and using VecMap to align the two vectors to the same vector space.

The results at Table~\ref{tabfast} and Table~\ref{tabword} show the coverage and Spearman results. Coverage refers to the total number of in vocabulary words (words that are found both in the model and evaluation dataset). We can see that the coverage is lower for word2Vec but a little higher for fastText (we expected coverage for fastText to be 100 percent). The Simlex and WordSim similarity score for monolingual fastText embeddings in Table~\ref{tabfast} is higher, this is expected due to the coverage percentage also being very high as compared to the coverage value in Table~\ref{tabword}.


 \begin{table}[t]
 \caption{FastText Monolingual Results}\label{tabfast}
 \begin{tabular}{llll} 
 Monolingual fastText & Coverage & Spearman \\
  Sepedi(Simlex)	& 94.58 & 40.39 \\
  Sepedi(WordSim)	& 81.29 & 46.15 \\
  Setswana(Simlex) & 95.22 & 33.23\\
  Setswana(WordSim) & 95.38 & 44.80\\
 \end{tabular}
 \end{table}

 \begin{table}[t]
 \caption{Word2Vec Monolingual Results}\label{tabword}
 \begin{tabular}{llll} 
 Monolingual word2Vec & Coverage & Spearman  \\
   Sepedi(Simlex)	& 79.49 & 25.96 \\
   Sepedi(WordSim)	& 84.49  & 23.57 \\
   Setswana(Simlex) & 95.32 & 31.52  \\
   Setswana(WordSim) & 95.38 & 35.11 &\\
 \end{tabular}
 \end{table}

 \begin{table}[t]
 \caption{Word2Vec Crosslingual Results}\label{tabcrossw}
 \begin{tabular}{llll} 
 Monolingual word2Vec & Coverage & Spearman\\
 Setswana-Sepedi(Simlex) & 90.76 & 31.14 \\
 Setswana-Sepedi(WordSim) & 68.56 & 40.87 \\
 \end{tabular}
 \end{table}


 \begin{table}[t]
 \caption{FastText Crosslingual Results}\label{tabcrossf}
 \begin{tabular}{llll} 
 Crosslingual fastText & Coverage & Spearman\\
 Setswana-Sepedi(Simlex) & 91.19 & 30.44\\
 Setswana-Sepedi(WordSim) & 68.84 & 36.33\
 \end{tabular}
 \end{table}

\section{Discussion}\label{sec:disc}
The main purpose of this research is to show that it is possible to do cross-lingual transfer from the source language to the target. In essence we wanted to check if cross-lingual alignment can improve the word representation for the target language. The results on Table~\ref{tabword} shows that the Spearman's correlation value for the target language when using word2Vec is low, this is also due to coverage percentage, but fastText based-embeddings perform better on Table~\ref{tabfast}  and has a higher coverage percentage, as stated upove we expected 100 percent coverage. Table~\ref{tabcrossw} shows that we improved the representation of words after cross-lingual alignment for word2Vec based-embeddings. The Spearman's value has increased for both Simlex and Wordsim. We expected to improve the results for fastText embeddings but in this case word2Vec actually yielded better results.

\section{Conclusion}\label{sec:con}
In this paper, VecMap was used to align Setswana-Sepedi to the same vector space. Through this work, we wanted to use cross-lingual (VecMap) technique to enable knowledge transfer between languages with rich resources and low resources. The results show that it is possible to align two monolingual embeddings to get cross-lingual embeddings. We mapped Setswana to Sepedi and used Spearman's to check correlation. Interestingly we get different results on fastText and word2Vec-based embeddings though we used the same data to train the embeddings.

In future work, it would be interesting to use the cross-lingual embedding on a downstream task like translation or sentiment analysis specifically for low-resourced languages.

\section{Acknowledgements}

We would like to acknowledge ABSA for sponsoring the industry chair and it’s related activities to the project.

\bibliographystyle{agsm}

\bibliography{bibliography.bib}

@article{artetxe2019massively,
	author = {Artetxe, Mikel and Schwenk, Holger},
	journal = {Transactions of the Association for Computational Linguistics},
	pages = {597--610},
	publisher = {MIT Press},
	title = {Massively multilingual sentence embeddings for zero-shot cross-lingual transfer and beyond},
	volume = {7},
	year = {2019}}

@inproceedings{adams2017cross,
	author = {Adams, Oliver and Makarucha, Adam and Neubig, Graham and Bird, Steven and Cohn, Trevor},
	booktitle = {Proceedings of the 15th Conference of the European Chapter of the Association for Computational Linguistics: Volume 1, Long Papers},
	pages = {937--947},
	title = {Cross-lingual word embeddings for low-resource language modeling},
	year = {2017}}

@inproceedings{ruder2019transfer,
	author = {Ruder, Sebastian and Peters, Matthew E and Swayamdipta, Swabha and Wolf, Thomas},
	booktitle = {Proceedings of the 2019 Conference of the North American Chapter of the Association for Computational Linguistics: Tutorials},
	pages = {15--18},
	title = {Transfer learning in natural language processing},
	year = {2019}}

@inproceedings{hedderich2020transfer,
	author = {Hedderich, Michael A and Adelani, David and Zhu, Dawei and Alabi, Jesujoba and Markus, Udia and Klakow, Dietrich},
	booktitle = {Proceedings of the 2020 Conference on Empirical Methods in Natural Language Processing (EMNLP)},
	pages = {2580--2591},
	title = {Transfer Learning and Distant Supervision for Multilingual Transformer Models: A Study on African Languages},
	year = {2020}}

@inproceedings{marivate2020improving,
	author = {Marivate, Vukosi and Sefara, Tshephisho},
	booktitle = {International Cross-Domain Conference for Machine Learning and Knowledge Extraction},
	organization = {Springer},
	pages = {385--399},
	title = {Improving short text classification through global augmentation methods},
	year = {2020}}

@inproceedings{marivate2020investigating,
  title={Investigating an approach for low resource language dataset creation, curation and classification: Setswana and Sepedi},
  author={Marivate, Vukosi and Sefara, Tshephisho and Chabalala, Vongani and Makhaya, Keamogetswe and Mokgonyane, Tumisho and Mokoena, Rethabile and Modupe, Abiodun},
  booktitle={Proceedings of the first workshop on Resources for African Indigenous Languages},
  pages={15--20},
  year={2020}
 }

@article{bakarov2018survey,
	author = {Bakarov, Amir},
	journal = {arXiv preprint arXiv:1801.09536},
	title = {A survey of word embeddings evaluation methods},
	year = {2018}}

@article{alabi2019massive,
	author = {Alabi, Jesujoba O and Amponsah-Kaakyire, Kwabena and Adelani, David I and Espa{\~n}a-Bonet, Cristina},
	journal = {arXiv preprint arXiv:1912.02481},
	title = {Massive vs. Curated Word Embeddings for Low-Resourced Languages. The Case of Yor$\backslash$ub$\backslash$'a and Twi},
	year = {2019}}

@article{bojanowski2016enriching,
	author = {Bojanowski, Piotr and Grave, Edouard and Joulin, Armand and Mikolov, Tomas},
	journal = {arXiv preprint arXiv:1607.04606},
	title = {Enriching Word Vectors with Subword Information},
	year = {2016}}

@article{hill2015simlex,
	author = {Hill, Felix and Reichart, Roi and Korhonen, Anna},
	journal = {Computational Linguistics},
	number = {4},
	pages = {665--695},
	publisher = {MIT Press One Rogers Street, Cambridge, MA 02142-1209, USA journals-info~{\ldots}},
	title = {Simlex-999: Evaluating semantic models with (genuine) similarity estimation},
	volume = {41},
	year = {2015}}

@inproceedings{finkelstein2001placing,
	author = {Finkelstein, Lev and Gabrilovich, Evgeniy and Matias, Yossi and Rivlin, Ehud and Solan, Zach and Wolfman, Gadi and Ruppin, Eytan},
	booktitle = {Proceedings of the 10th international conference on World Wide Web},
	pages = {406--414},
	title = {Placing search in context: The concept revisited},
	year = {2001}}

@inproceedings{artetxe2018robust,
	author = {Artetxe, Mikel and Labaka, Gorka and Agirre, Eneko},
	booktitle = {Proceedings of the 56th Annual Meeting of the Association for Computational Linguistics (Volume 1: Long Papers)},
	pages = {789--798},
	title = {A robust self-learning method for fully unsupervised cross-lingual mappings of word embeddings},
	year = {2018}}

@inproceedings{lample2018word,
	author = {Lample, Guillaume and Conneau, Alexis and Ranzato, Marc'Aurelio and Denoyer, Ludovic and J{\'e}gou, Herv{\'e}},
	booktitle = {International Conference on Learning Representations},
	title = {Word translation without parallel data},
	year = {2018}}

@inproceedings{guo2014revisiting,
	author = {Guo, Jiang and Che, Wanxiang and Wang, Haifeng and Liu, Ting},
	booktitle = {Proceedings of the 2014 Conference on Empirical Methods in Natural Language Processing (EMNLP)},
	pages = {110--120},
	title = {Revisiting embedding features for simple semi-supervised learning},
	year = {2014}}

@inproceedings{socher2013recursive,
	author = {Socher, Richard and Perelygin, Alex and Wu, Jean and Chuang, Jason and Manning, Christopher D and Ng, Andrew Y and Potts, Christopher},
	booktitle = {Proceedings of the 2013 conference on empirical methods in natural language processing},
	pages = {1631--1642},
	title = {Recursive deep models for semantic compositionality over a sentiment treebank},
	year = {2013}}

@inproceedings{mikolov2013distributed,
	author = {Mikolov, Tomas and Sutskever, Ilya and Chen, Kai and Corrado, Greg S and Dean, Jeff},
	booktitle = {Advances in neural information processing systems},
	pages = {3111--3119},
	title = {Distributed representations of words and phrases and their compositionality},
	year = {2013}}

@inproceedings{artetxe2020cross,
	author = {Artetxe, Mikel and Ruder, Sebastian and Yogatama, Dani},
	booktitle = {Proceedings of the 58th Annual Meeting of the Association for Computational Linguistics},
	pages = {4623--4637},
	title = {On the Cross-lingual Transferability of Monolingual Representations},
	year = {2020}}

@inproceedings{howard2018universal,
	author = {Howard, Jeremy and Ruder, Sebastian},
	booktitle = {Proceedings of the 56th Annual Meeting of the Association for Computational Linguistics (Volume 1: Long Papers)},
	pages = {328--339},
	title = {Universal Language Model Fine-tuning for Text Classification},
	year = {2018}}

@article{kwiatkowski2019natural,
	author = {Kwiatkowski, Tom and Palomaki, Jennimaria and Redfield, Olivia and Collins, Michael and Parikh, Ankur and Alberti, Chris and Epstein, Danielle and Polosukhin, Illia and Devlin, Jacob and Lee, Kenton and others},
	journal = {Transactions of the Association for Computational Linguistics},
	pages = {453--466},
	publisher = {MIT Press},
	title = {Natural questions: a benchmark for question answering research},
	volume = {7},
	year = {2019}}

@article{devlin2018bert,
	author = {Devlin, Jacob and Chang, Ming-Wei and Lee, Kenton and Toutanova, Kristina},
	journal = {arXiv preprint arXiv:1810.04805},
	title = {Bert: Pre-training of deep bidirectional transformers for language understanding},
	year = {2018}}

@article{cho2014learning,
	author = {Cho, Kyunghyun and Van Merri{\"e}nboer, Bart and Gulcehre, Caglar and Bahdanau, Dzmitry and Bougares, Fethi and Schwenk, Holger and Bengio, Yoshua},
	journal = {arXiv preprint arXiv:1406.1078},
	title = {Learning phrase representations using RNN encoder-decoder for statistical machine translation},
	year = {2014}}

@inproceedings{nekoto2020participatory,
	author = {Nekoto, Wilhelmina and Marivate, Vukosi and Matsila, Tshinondiwa and Fasubaa, Timi and Fagbohungbe, Taiwo and Akinola, Solomon Oluwole and Muhammad, Shamsuddeen and Kabenamualu, Salomon Kabongo and Osei, Salomey and Sackey, Freshia and others},
	booktitle = {Proceedings of the 2020 Conference on Empirical Methods in Natural Language Processing: Findings},
	pages = {2144--2160},
	title = {Participatory Research for Low-resourced Machine Translation: A Case Study in African Languages},
	year = {2020}}

@article{martinus2019focus,
	author = {Martinus, Laura and Abbott, Jade Z},
	journal = {arXiv preprint arXiv:1906.05685},
	title = {A focus on neural machine translation for african languages},
	year = {2019}}

@inproceedings{abbott2019benchmarking,
	author = {Abbott, Jade and Martinus, Laura},
	booktitle = {Proceedings of the 2019 Workshop on Widening NLP},
	pages = {98--101},
	title = {Benchmarking neural machine translation for southern African languages},
	year = {2019}}

@article{ranathunga2021neural,
	author = {Ranathunga, Surangika and Lee, En-Shiun Annie and Skenduli, Marjana Prifti and Shekhar, Ravi and Alam, Mehreen and Kaur, Rishemjit},
	journal = {arXiv preprint arXiv:2106.15115},
	title = {Neural machine translation for low-resource languages: A survey},
	year = {2021}}

@inproceedings{inproceedings,
    author = {Pennington, Jeffrey and Socher, Richard and Manning, Christopher},
    year = {2014},
    month = {01},
    pages = {1532-1543},
    title = {Glove: Global Vectors for Word Representation},
    volume = {14},
    journal = {EMNLP},
    doi = {10.3115/v1/D14-1162}
}

@article{Hochreiter1997LongSM,
  title={Long Short-Term Memory},
  author={Sepp Hochreiter and J{\"u}rgen Schmidhuber},
  journal={Neural Computation},
  year={1997},
  volume={9},
  pages={1735-1780}
}

@inproceedings{agic-vulic-2019-jw300,
    title = "{JW}300: A Wide-Coverage Parallel Corpus for Low-Resource Languages",
    author = "Agi{\'c}, {\v{Z}}eljko  and
      Vuli{\'c}, Ivan",
    booktitle = "Proceedings of the 57th Annual Meeting of the Association for Computational Linguistics",
    month = jul,
    year = "2019",
    address = "Florence, Italy",
    publisher = "Association for Computational Linguistics",
    url = "https://aclanthology.org/P19-1310",
    doi = "10.18653/v1/P19-1310",
    pages = "3204--3210",
}

@misc{marivate_vukosi_2020_3668495,
  author       = {Marivate, Vukosi and
                  Sefara, Tshephisho},
  title        = {South African News Data},
  month        = feb,
  year         = 2020,
  publisher    = {Zenodo},
  version      = {0.5.1},
  doi          = {10.5281/zenodo.3668495},
  url          = {https://doi.org/10.5281/zenodo.3668495}
}

@inproceedings{eiselen2014developing,
  title={Developing Text Resources for Ten South African Languages.},
  author={Eiselen, Roald and Puttkammer, Martin J},
  booktitle={LREC},
  pages={3698--3703},
  year={2014}
}

@misc{conneau2018word,
      title={Word Translation Without Parallel Data}, 
      author={Alexis Conneau and Guillaume Lample and Marc'Aurelio Ranzato and Ludovic Denoyer and Hervé Jégou},
      year={2018},
      eprint={1710.04087},
      archivePrefix={arXiv},
      primaryClass={cs.CL}
}

@misc{dinu2015improving,
      title={Improving zero-shot learning by mitigating the hubness problem}, 
      author={Georgiana Dinu and Angeliki Lazaridou and Marco Baroni},
      year={2015},
      eprint={1412.6568},
      archivePrefix={arXiv},
      primaryClass={cs.CL}
}

@inproceedings{sefara2021transformer,
  title={Transformer-based Machine Translation for Low-resourced Languages embedded with Language Identification},
  author={Sefara, Tshephisho J and Zwane, Skhumbuzo G and Gama, Nelisiwe and Sibisi, Hlawulani and Senoamadi, Phillemon N and Marivate, Vukosi},
  booktitle={2021 Conference on Information Communications Technology and Society (ICTAS)},
  pages={127--132},
  year={2021},
  organization={IEEE}
}

@inproceedings{ijcai2018-566,
  title     = {Improving Low Resource Named Entity Recognition using Cross-lingual Knowledge Transfer},
  author    = {Xiaocheng Feng and Xiachong Feng and Bing Qin and Zhangyin Feng and Ting Liu},
  booktitle = {Proceedings of the Twenty-Seventh International Joint Conference on
               Artificial Intelligence, {IJCAI-18}},
  publisher = {International Joint Conferences on Artificial Intelligence Organization},             
  pages     = {4071--4077},
  year      = {2018},
  month     = {7},
  doi       = {10.24963/ijcai.2018/566},
  url       = {https://doi.org/10.24963/ijcai.2018/566},
}

@misc{banerjee2021crosslingual,
      title={Crosslingual Embeddings are Essential in UNMT for Distant Languages: An English to IndoAryan Case Study}, 
      author={Tamali Banerjee and Rudra Murthy V au2 and Pushpak Bhattacharyya},
      year={2021},
      eprint={2106.04995},
      archivePrefix={arXiv},
      primaryClass={cs.CL}
}

\end{document}